%% file: main.tex
\documentclass[sigconf, nonacm]{acmart}
\AtBeginDocument{%
  \providecommand\BibTeX{{%
    \normalfont B\kern-0.5em{\scshape i\kern-0.25em b}\kern-0.8em\TeX}}}

\copyrightyear{2023}
\acmYear{2023}
\setcopyright{rightsretained}
\acmConference[EAAMO '23]{Equity and Access in Algorithms, Mechanisms, and Optimization}{October 30-November 1, 2023}{Boston, MA, USA}
\acmBooktitle{Equity and Access in Algorithms, Mechanisms, and Optimization (EAAMO '23), October 30-November 1, 2023, Boston, MA, USA}
\acmDOI{10.1145/3617694.3623251}
\acmISBN{979-8-4007-0381-2/23/11}

\usepackage{graphicx}  
\usepackage{amsmath}
\usepackage{soul}  
\usepackage{hyperref}
\usepackage{tabularx}
\usepackage{xspace}
\usepackage{algorithm}
\usepackage[noend]{algpseudocode}
\usepackage{circledsteps}
\usepackage{courier}

\pgfkeys{/csteps/inner color=white}
\pgfkeys{/csteps/outer color=black}
\pgfkeys{/csteps/fill color=black}

\newcommand{\etal}{et al.\xspace}
\newcommand*{\eg}{e.g.,\xspace}
\newcommand*{\ie}{i.e.,\xspace}

\author{Joao Fonseca}
\authornote{Both authors contributed equally}
\affiliation{%
 \institution{NOVA University Lisbon}
  \city{Lisbon}
  \country{Portugal}
}
\email{jpfonseca@novaims.unl.pt}

\author{Andrew Bell}
\authornotemark[1]
\affiliation{%
 \institution{New York University}
  \city{New York, NY}
  \country{USA}
}
\email{alb9742@nyu.edu}
\author{Carlo Abrate}
\affiliation{%
 \institution{CENTAI}
  \city{Turin}
  \country{Italy}
}
\email{carlo.abrate@centai.eu}
\author{Francesco Bonchi}
\affiliation{%
 \institution{CENTAI, Turin, Italy}
 \institution{Eurecat, Barcelona, Spain}
 \country{}
}
\email{bonchi@centai.eu}
\author{Julia Stoyanovich}
\affiliation{%
 \institution{New York University}
  \city{New York, NY}
  \country{USA}
}
\email{stoyanovich@nyu.edu}

\begin{document}

\title{Setting the Right Expectations: Algorithmic Recourse Over Time}

\begin{abstract}
\input{abstract}
\end{abstract}

\maketitle

\input{introduction}
\input{related}
\input{problem}
\input{experiments}

\input{results}

\input{discussion}
\input{conc}

\input{ack}

\bibliographystyle{ACM-Reference-Format}
\bibliography{references}

\end{document}

%% file: abstract.tex
Algorithmic systems are often called upon to assist in high-stakes decision making.  In light of this, \emph{algorithmic recourse}, the principle wherein individuals should be able to take action against an undesirable outcome made by an algorithmic system, is receiving growing attention. The bulk of the literature on algorithmic recourse to-date focuses primarily on how to provide recourse to a single individual, overlooking a critical element: the effects of a continuously changing context. Disregarding these effects on recourse is a significant oversight, since, in almost all cases, recourse consists of an individual making a first, unfavorable attempt, and then being given an opportunity to make one or several attempts \emph{at a later date} --- when the context might have changed. This can create false expectations, as initial recourse recommendations may become less reliable over time due to model drift and competition for access to the favorable outcome between individuals.

In this work we propose an agent-based simulation framework for studying the effects of a continuously changing environment on algorithmic recourse. In particular, we identify two main effects that can alter the \emph{reliability of recourse} for individuals represented by the agents: (1) competition with other agents acting upon recourse, and (2) competition with new agents entering the environment. Our findings highlight that only a small set of specific parameterizations result in algorithmic recourse that is reliable for agents over time. Consequently, we argue that substantial additional work is needed to understand recourse reliability over time, and to develop recourse methods that reward agents' effort.

%% file: introduction.tex
\section{Introduction}
\label{sec:introduction}

\begin{figure}[t!]
    \centering
    \includegraphics[width=\linewidth]{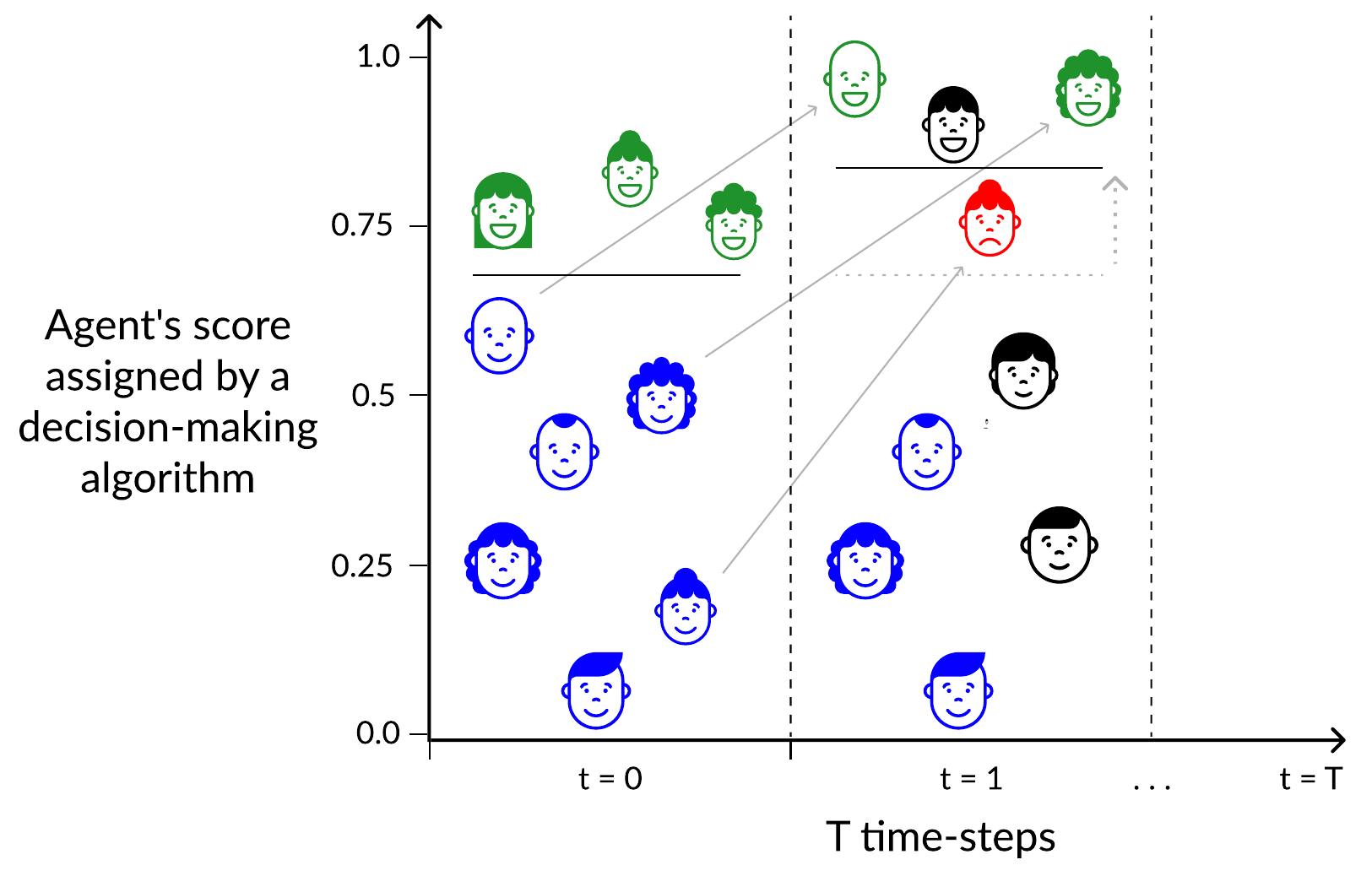}
    \caption{Motivation for multi-agent over-time analysis of algorithmic recourse. First, at time $t=0$, blue agents received a negative outcome and green agents received a positive outcome. Second, at time $t=1$, black agents entered the environment. The red agent that had the negative outcome at $t=0$ took actions according to the recourse recommendation generated at $t=0$, but (disappointingly) such recourse-guided effort turned out to be insufficient. This is because context has changed from $t=0$ to $t=1$: competition from other agents acting upon recourse and from new agents ``raised the bar,'' pushing up the decision boundary.}
    \label{fig:sketch}
\end{figure}

Artificial intelligence (AI) systems are becoming increasingly common in consequential decision-making settings such as healthcare \cite{begoli2019need, grote2020ethics}, finance \cite{mukerjee2002multi},  and hiring \cite{schumann2020we, raghavan2020mitigating, karimi2022survey}. While these systems have the capacity to significantly improve people's lives, they can also have adverse consequences, such as erroneous decision-making~\cite{steimers2022sources}. As a result, those developing and researching AI systems in high-stakes domains have introduced the concept of \emph{algorithmic recourse}, which is the ability of an individual to take action against the outcome of an algorithmic decision-making system. Algorithmic recourse allows an agent (\ie individual) to understand:
\begin{enumerate}
  \item \textit{why} an outcome was produced by the system, and
  \item \textit{what} can be done in order to reverse it \cite{karimi2021algorithmic, ustun2019actionable}.
\end{enumerate}

The importance of algorithmic recourse has been argued from an equitable and ethical computing standpoint ~\cite{venkatasubramanian2020philosophical, schermer2011limits}, and significant effort has been devoted to the \emph{when}, \emph{why}, and \emph{how} of giving recourse to individuals. However, as we observe in Section~\ref{sec:related}, time-related effects, producing a continuously changing context, have been considered only by a few authors. \textbf{The driving insight of this paper is that the impact of time and of a continuously changing context should never be ignored in algorithmic recourse work because \emph{time is intrinsic to the notion of recourse itself.}}  Typically, recourse consists of an individual making a first, unsuccessful, attempt at a time $t$, and then being given an opportunity to make a second attempt at a \emph{later time} $t + \delta$. Depending on the length of time represented by $\delta$, recourse recommendations from $t$ may become demonstrably less reliable. The importance of considering time-related effects has been acknowledged by Ferrario and Loi, and Barocas~\etal, further highlighting the importance of continued work aimed at closing this critical research gap~\cite{ferrario2022robustness,barocas2020hidden}.

As an example, consider the lending setting, wherein an AI system denies an individual's application for a loan but provides information on what that individual can do to be approved for the loan \emph{if they apply again at a later date}~\cite{howarth2016comparative}. The individual may be told that their loan application was denied because their credit score is 50 points lower than necessary. 
One could imagine that it takes the individual 6 months to a year to improve their credit score --- which is enough time for the criteria for approving the loan to change. As a result, the initial recommendation of \emph{``improving your credit score by 50 points''} may have set false expectations.

There are numerous reasons why selection criteria --- and the reliability of recourse recommendations --- can change over time~\cite{o2021multi,rawal2020beyond,sullivan2022explanation,upadhyay2021towards}. In this work, we look at competitive effects that arise from having a multi-agent resource-constrained setting. In particular, we identify two main effects: 
(1) other agents are acting upon recourse recommendations they have received, and (2) new agents are entering the system. An illustration of the competitive effects and their impact on recourse can be seen in Figure~\ref{fig:sketch}.

\begin{figure}[t!]
    \centering
    \includegraphics[width=1.05\linewidth]{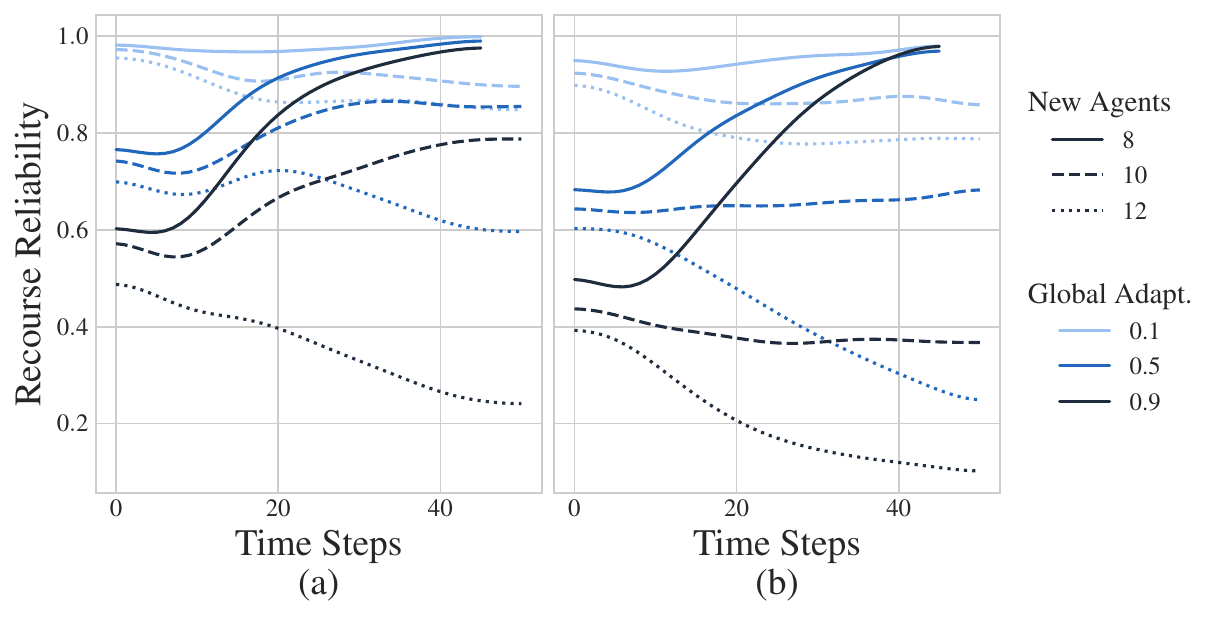}
    \caption{Comparison of recourse reliability (metric defined in Section \ref{sec:problem:metrics}) over time under different simulated models of agent behavior. Based on the behaviors defined in Section~\ref{sec:problem:simulation}, (a) shows simulations where agents' behavior is modeled under continuous adaptation with constant effort; (b) shows simulations under continuous adaptation with flexible effort.}
    \label{fig:success-rate-comparison}
\end{figure}

\smallskip
\noindent \textbf{Research questions, contributions, and roadmap.} Motivated by the scenarios described above, this work aims to develop a framework for multi-agent multi-time-step analysis of algorithmic recourse. We seek to answer the following research questions:
\begin{enumerate}
    \item What metrics can be used to evaluate the reliability of algorithmic recourse over time?
    \item Under what conditions is algorithmic recourse most reliable for individuals?
    \item How can system designers set better expectations for what will happen to individuals, if they follow algorithmic recourse recommendations?
\end{enumerate}

We tackle these questions with the help of an agent-based simulation framework, in which a population of agents, each having its own features, applies for access to a scarce resource, and a black-box model decides on the outcomes.
Agents obtaining a positive outcome exit the system, while those obtaining a negative outcome are given recourse recommendations, allowing them to adapt their features before applying again at the next time-step. Our framework is guided by a rich set of stochastic variables, including how flexible/willing  an agent is to adapt and how many new agents are joining the system at each time step to compete for limited resources. To the best of our knowledge, this is the only work examining algorithmic recourse under resource-constrained settings using agent-based modeling. We are also among the first authors to study recourse over many time steps, rather than just examining one or two points in time.

Moreover, we define a \emph{recourse reliability} metric, quantifying the probability that individuals will receive a favorable outcome, given that they took the recommended recourse. We also use our framework to assess how different parameterizations of the setting affect this metric.  Our findings highlight that only a small set of specific parameterizations results in algorithmic recourse that is reliable for individuals over time. As a preview of our results, Figure~\ref{fig:success-rate-comparison} shows that, under 9 different parameterizations of our framework, one rarely observes reliable recourse over 50 time-steps.

Our work highlights a crucial socio-technical implication: Systems that provide recourse recommendations \emph{without considering potential shifts in the threshold for positive outcomes} can create unrealistic expectations for individuals. In some cases, these recommendations may even be harmful by falsely promising rewards to those who seek recourse based on recommendations.  This calls for substantial additional work on understanding recourse reliability in multi-agent multi-timestep settings, and on developing recourse methods that reward agents' effort.

The rest of this paper is organized as follows:
\begin{itemize}
  \item We review related work on algorithmic recourse (Section~\ref{sec:related}).
  \item We formalize multi-agent algorithmic recourse, propose a simulation framework for evaluating recourse over time, and define a recourse reliability metric (Section~\ref{sec:problem}).
  \item We present detailed analysis that confirms the importance of considering temporal effects in recourse (Section~\ref{sec:methods}).
  \item We provide an in-depth discussion of the practical implications of our observations (Section~\ref{sec:disc}).
\end{itemize}

%% file: related.tex
\section{Related Work}
\label{sec:related}

Algorithmic recourse is critically important for three reasons: first, as mentioned, it ascribes agency to individuals against adverse outcomes, including outcomes that are either incorrect (and inefficient) or discriminatory~\cite{ustun2019actionable, von2022fairness, sullivan2022explanation}. As such, many have argued that providing individuals recourse is morally good and equitable, and therefore consistent with ethical computing~\cite{venkatasubramanian2020philosophical, schermer2011limits}. Second, from the perspective of the owners of AI systems, recourse can improve the overall accuracy and reliability of a system's outcomes. Third, algorithmic recourse will likely become legally necessary with the passing of legislation like the European Union AI Act~\cite{bell2023think, cordeiro2019civil}.

There are two main approaches used for providing algorithmic recourse in practice. The first and simpler approach is to provide recourse based on ``contrastive explanations,'' which find changes that can be made to an individual's profile (\ie the \emph{feature space}) to flip their outcome from unfavorable to favorable~\cite{ustun2019actionable, bottou2013counterfactual}. These recommended changes are sometimes called ``interventions.'' While useful in many settings, a short-coming of this approach is that it does not consider the normative meaning of features. This weakness can lead to recourse recommendations that are disconnected from real-world actions~\cite{barocas2020hidden, kumar2020problems}, like recommending to an individual that they should ``lower their age`` to receive a different outcome.

The second set of methods, ``causal recourse methods,'' use structural causal models~\cite{pearl2000models, karimi2020algorithmic} to account for downstream effects when changing a smaller set of an individual's features (also referred to as ``consequential recommendations''~\cite{karimi2020survey}). The advantage of causal recourse methods is that they produce smaller, easier to achieve intervention sets for individuals~\cite{karimi2021algorithmic} that are better connected with real-world actions and settings.

\subsection{Temporal effects on recourse}
\label{sec:related:temporal}

To the best of our knowledge, the impact of time in algorithmic recourse is understudied. As noted earlier, we see this as a significant research gap, because \emph{time is inherent to algorithmic recourse itself}. We now discuss several studies that begin to fill this gap.

Ferrario and Loi, motivated by the idea that machine learning models are often unstable in real-world settings, studied the impact of retraining models on the validity of counterfactual explanations over time~\cite{ferrario2022robustness}. In their work, they developed a method using counterfactual data augmentation to improve the robustness of recourse recommendations and prevent so-called \emph{Unfortunate Counterfacutal Events (UCEs)}, which occur when an individual is given a recourse recommendation at time-step $t$, but, due to model updates, the recommendation is invalid at time-step $t+\delta$. Other researchers have more generally studied the robustness of counterfactual explanations due to changes in the underlying settings, but, notably, they did not emphasize temporal effects as the reason for these changes~\cite{ehyaei2023robustness, guyomard2023generating, mishra2021survey, virgolin2023robustness, upadhyay2021towards}.

Rawal~\etal analyzed the impact of distribution shifts in the data on algorithmic recourse recommendations~\cite{rawal2020algorithmic}. They did not make any assumptions about the origin of these distribution shifts, and considered temporal shifts, geo-spatial shifts, and shifts due to data correction. An important finding of this work is that there are theoretical trade-offs between minimizing the cost of a recourse recommendations and ensuring robustness to distribution shifts. Under this trade-off, Pawelczyk~\etal proposed a method for generating counterfactual explanations that allows individuals to set preferences and navigate this trade-off for themselves~\cite{pawelczyk2022probabilistically}. We further refer to the work of Pawelczyk~\etal, and Ferrario and Loi in Section~\ref{sec:problem:metrics}.

The methods by \citet{upadhyay2021towards} and \citet{rawal2020beyond} use adversarial training to provide recourse recommendations that are robust to model shifts. They quantify the probability of recourse being invalidated given the possibility of model shifts. This is achieved by defining a set of plausible model shifts based on perturbations introduced in the parameter or gradient space. This analysis applied multiple data drift scenarios over two time steps.

Our work fills an important gap left by existing work: Rather than focusing specifically on the robustness of counterfactual explanations, we look to characterize the reliability of algorithmic recourse more generally. The framework we propose is agnostic to the method used to generate recourse recommendations. Furthermore, we not only consider temporal effects in algorithmic recourse, but also study them alongside multi-agent effects, and in resource-constrained environments.

\subsection{Recourse in multi-agent settings}
\label{sec:related:miltiagent}

The vast majority of work on both contrastive explanations and causal recourse methods has focused on single-agent settings, save for a few exceptions~\cite{upadhyay2021towards, rawal2020beyond, o2021multi, o2022toward, altmeyer2023endogenous}. One notable work on multi-agent algorithmic recourse is by
\citet{o2021multi, o2022toward}, who adapted concepts from game theory literature to algorithmic recourse. They defined metrics like ``Social-Welfare-Efficient recourse'' and ``Pareto-efficient recourse,'' which measure the effect of one agent taking recourse on the population of agents as a whole. Their analysis of multi-agent interactions, done through the lens of the prisoners' dilemma problem, leads to the following key conclusions: The improvement of an individual (or subgroup) tends to result in a loss in social welfare, while no scenario results in an improvement of both social welfare and the principal agent's. This finding highlights an open question: ``When should algorithmic recourse be provided?'' and is related to insights from \citet{barocas2020hidden}, who suggest that recourse should not be presented to someone if it encourages an action that would be harmful to them.

Another relevant line of work is by \citet{altmeyer2023endogenous}, who studied how recourse-based multi-agent interactions affect model drift over multiple time-steps. Their analysis shows significant model drift effects, which changed the threshold for receiving a positive outcome over time. This work has similarities to our own, however, their conclusions only hold under a restricted set of assumptions, namely, that (1) an agent who follows recourse recommendations is guaranteed a positive outcome; (2) the model is retrained at every time-step, using all prior data; and (3) agents are capable of taking as much recourse-based action as they need, and make the exact changes recommended to them. Importantly, violations of the first assumption could lead to unrealistic outcomes (\eg decreasing an agent's age), or, in the case of the loan example, to significant monetary losses \cite{upadhyay2021towards}.

Our contribution departs from \citet{o2021multi}, and \citet{altmeyer2023endogenous}, as we propose a more realistic agent-based framework, allowing agents to act on recourse in different ways, and accommodating more flexible population dynamics, with new agents joining and winning agents exiting the system. There are two other distinctions present in our simulation framework. First, at each iteration, we provide recourse to all agents who did not receive a favorable outcome, and they take recourse action based on a function describing their actions (also potentially choosing not to act). In contrast, in \citet{altmeyer2023endogenous}, a batch of agents is selected and provided recourse. Second, \citet{altmeyer2023endogenous} focus on gradient-based counterfactual search, while our formulation is agnostic to the counterfactual search method.

%% file: problem.tex
\section{Proposed Framework}
\label{sec:problem}

We study the problem of recourse in binary classification, where receiving the positive outcome corresponds to an agent gaining access to a desirable resource. Note that from here forward we use ``agent'' to mean an individual who is receiving an outcome from a system and possibly taking recourse. We start from the well-studied recourse problem in the static setting for a single agent, and then move to a more realistic setting in which multiple agents are competing for access to scarce resources (\ie the number of agents competing exceeds the number of available positive outcomes). \textbf{Our goal in developing this framework was to create a way to realistically simulate algorithmic recourse in a multi-agent, multi-time step environment.} 

\subsection{Recourse for a Single Agent}
\label{sec:problem:recourse_single}

Consider a single agent, described by a set of features, $x \in \mathcal{X}$, and a black-box classifier $f: \mathcal{X} \rightarrow \{0,1\}$ that is used to generate an outcome for the agent.  If an agent $x$ receives a negative outcome from the classifier $f$, then the outcome is supplemented with a recommendation (or an explanation) on what they can change in their feature space to receive a positive outcome instead. Generating this recommendation is the goal of the single-agent recourse problem.

\begin{definition}[Single-agent recourse problem] \label{def:rec}
Given a black-box classifier $f: \mathcal{X} \rightarrow \{0,1\}$ and an agent $x \in \mathcal{X}$ for which $f(x)=0$ (a negative outcome), find a new configuration of the agent $x' \in \mathcal{X}$ that results in $f(x') = 1$ (a positive outcome), and is associated with the lowest cost $c(x,x')$ of making the change:
\begin{equation}
\begin{split}
x' = \min_{x'} \quad & c(x,x') \\
s.t. \quad & f(x') = 1 \\
& x' \in \mathcal{X}
\end{split}
\end{equation}

Use the configuration $x'$ to return a recommendation to agent $x$ regarding the adjustment they need to make to their features to receive a positive outcome.
\end{definition}

To make this definition more concrete, consider a scenario where agents are applying for a loan form a bank (a common example used when discussing recourse). For our purposes, let an agent $x \in \mathcal{X}$ have two features: its credit rating $x_1$ and annual income $x_2$. Suppose the agent applies for the loan, but is ultimately denied based on the output of a risk assessment tool, represented by $f: \mathcal{X} \rightarrow \{0,1\}$.  If the bank decides to offer recourse to $x$, they may suggest that credit rating $x_1$ should be improved by $\delta_1$ or income $x_2$ be improved by  $\delta_2$, or both, minimizing cost $c: \mathcal{X}\times \mathcal{X} \rightarrow [0,1]$ for the agent, so that $f(x')=1$. Various methods for solving the problem of Definition~\ref{def:rec} have been proposed ~\cite{ustun2019actionable,bottou2013counterfactual,karimi2020survey}.

\subsection{Multi-Agent Recourse}
\label{sec:problem:recourse}
Suppose next that agents belong to a population $P$, and that they all are competing for access to a scarce resource. For example, there may be $|P| = N$ loan applicants, but the bank is only able to lend money to $k \ll N$ of them.

Because of the existence of the resource constraint $k$, there is no fixed threshold that can be used to determine whether the output of $f(x)$ corresponds to a positive or a negative outcome. Instead, we construct the black-box model so that, rather than returning a binary outcome, it returns a score: $f(x): \mathcal{X} \rightarrow  [0,1]$. This score is used to rank agents, from highest to lowest, and then selecting the top-$k$ agents $P^k \subseteq P$ to receive the positive outcome. While there is \emph{no fixed threshold} associated with the positive outcome, we can use the $k^{th}$ highest score $s$ as a threshold for receiving the positive outcome.  We write this formally below:

\begin{definition}[Multi-agent recourse problem] \label{def:multi-rec}
Given a black box classifier $f: \mathcal{X} \rightarrow [0,1]$, a population of agents $P$, and a resource constraint $k$, compute the score threshold $s$ for receiving the positive outcome.

For each agent $x \in \mathcal{X}$ for which $f(x)<s$ (a negative outcome), find a new counterfactual configuration of the agent $x' \in \mathcal{X}$ that results in $f(x') \geq s$ (a positive outcome), and is associated with the lowest cost $c(x,x')$ of making the change:
\begin{equation}
\begin{split}
x' = \min_{x'} \quad & c(x,x') \\
s.t. \quad & f(x') \geq s \\
& x' \in \mathcal{X}
\end{split}
\end{equation}

Use the configuration $x'$ for each agent $x$ for which $f(x)<s$ to return a recommendation regarding the adjustments they need to make to their features to receive the positive outcome.
\end{definition}

\subsection{Modeling Agents' Behaviors}
\label{sec:problem:simulation}

Having shown how recourse can be modeled for multiple agents, we now turn our attention to modeling the behaviour of a population of agents over time. We consider two important, realistic considerations about agents' behavior with respect to recourse:

\begin{enumerate}
    \item \emph{How faithfully an agent follows the recourse recommendation.} Agents may follow the recourse recommendation \emph{exactly}, or they may outperform (or underperform) the recommendation. For example, returning to the loan example described in Section~\ref{sec:problem:recourse_single}, one could imagine that if an agent is told to increase their credit score by 50 points, they may do so exactly, or they may actually increase their score by 40 point, or by 60 points. We call this consideration \textbf{``adaptation.''}\footnote{Pawelcyzk~\etal use the terms \emph{prescribed recourse} and \emph{implemented recourse} to refer to a similar concept.}

    \item \emph{The likelihood of an agent to take any action.} An agent that receives a recourse recommendation may or may not act on it. The likelihood that an agent will attempt a recourse action is be determined by several factors, such their implicit willingness to make challenges and the amount of effort the action requires. In the loan example, if an agent is told to increase their credit score by 20 points, they may be more likely to make the effort as opposed to being told to increase it by 200 points. We call this consideration \textbf{``effort.''}
\end{enumerate}

\noindent \textbf{Adaptation.}
The single-agent recourse problem (Definition~\ref{def:rec}) assumes that, when a recommendation for recourse is provided, an agent will change their features \emph{exactly according to the recommendation} they receive. Here, we relax this assumption and consider cases where there is uncertainty regarding whether an agent will, in fact, change their configuration in the way that the recommendation suggests.

We model an agent's actions in changing their features as a function $a(x,x'): \mathcal{X} \times \mathcal{X} \rightarrow \mathcal{X}$ that produces configuration $x''$, with the goal of reaching or exceeding the score threshold $s$. We consider three cases: (1) $x$ follows the suggested recourse recommendation exactly, \ie $a(x,x') = x'$ with $f(x') = s$; (2) $x$ declines to follow the recommendation, \ie $a(x,x') = x$, with $f(x) < s$; (3) $x$ changes their features in a way that is feasible but different than the recourse recommendation, \ie $a(x,x') = x''$ with score $f(x'')$ that may be greater than or less than $s$.

In our simulation framework, we model this behavior using the adaptation parameter that has two settings: \emph{binary} and \emph{continuous}. The \emph{binary} setting matches the classical view of recourse found in the literature, and assumes that all agents that take recourse action \emph{exactly match the recommendation} (their new feature configuration is $x'$). Under this setting, an agent acts on the recourse recommendation with some probability $p$ (\ie $a(x,x')=x'$), and it retains the original value of the features with probability $1-p$ (\ie $a(x,x')=x$). 

The \emph{continuous} setting accounts for what we believe is a more realistic modeling of behavior: that an agent makes progress towards a recourse recommendation, and may out- or under-perform the recommendation at any given time-step. In this setting, the agent produces configuration $x''$ according to some probabilistic model (\eg a Gaussian distribution). When taking actions, a value $\delta_x$ is sampled from this distribution, resulting in a new configuration $x''$, and a new score $f(x'') = f(x) + \delta_x$ is computed.

\begin{figure}[t!]
    \centering
    \includegraphics[width=.75\linewidth]{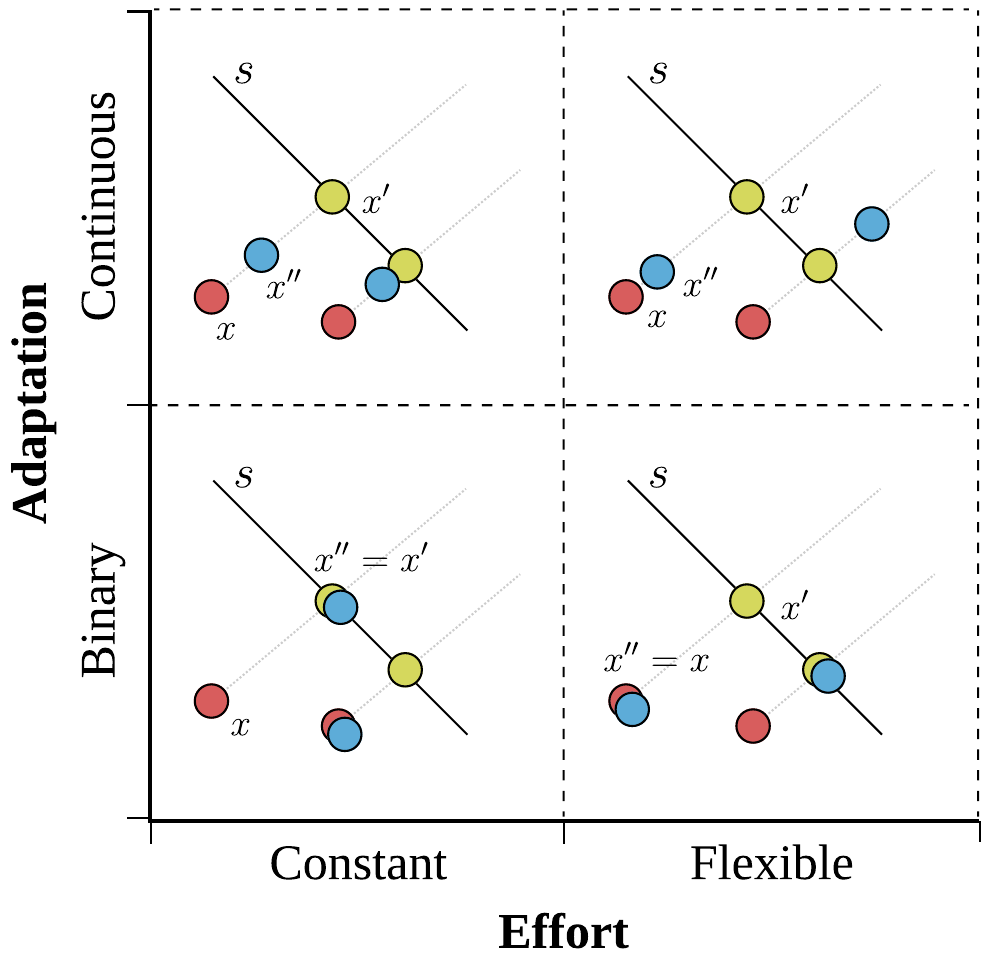}
    \caption{Different types of agent behavior. The agent (red) receives a recommendation (yellow) and takes action (blue). See Section~\ref{sec:problem:simulation} for a detailed description of the parameters \emph{adaptation} and \emph{effort}.}
    \label{fig:adaptation-vs-effort}
\end{figure}
 
\smallskip \noindent \textbf{Effort.} As described earlier, \emph{effort} reflects an agent's likelihood to take action and change their feature space. Under our simulation, this parameter has two settings: \emph{constant} and \emph{flexible.} The setting \emph{constant} means that an agent has an implicit willingness to act on recourse that is determined \emph{a priori} and is intrinsic to the agent. Regardless of what is happening in the environment, an agent has some probability $p$ that they will act on a recourse recommendation.

The setting \emph{flexible} means that the amount of effort required for a recourse recommendation determines the probability that an agent will take action and change their features. In other words, the less effort is required the more likely it is for an agent to act (and vice versa). In this case, the agents' probability to act on recourse is sampled from $\text{dist}(f(x),s)$. To encode effort into our simulation, we use the parameter $l \in \mathbb{R}$, and explain how how it is used in each setting in~\ref{sec:experiments:results}. $l$ is an additive factor, and intuitively, it can be thought of as the agent-level willingness to take recourse action.

\smallskip
\noindent \textbf{Difficulty of acting on recourse recommendations.} An additional consideration we use to model agent behavior is defining a global parameter that controls the \emph{difficulty of acting on a recourse recommendation}. For example, one can imagine that it is easier to act on a recourse recommendation when it is related to signing up for a social media account versus improving one's credit score.  The parameter $g \in [0,1]$ is set \emph{a priori} for the simulation. Values of $g$ closer to $1.0$ indicate a setting where it is easier for all agents to successfully get recourse.

\smallskip
\noindent \textbf{Adaptation and effort combined.} Since both adaption and effort have two settings, there are four possible settings to model agents' behavior. These are illustrated and described in Figure~\ref{fig:adaptation-vs-effort}. For better understanding, let's consider two of the settings in detail: 
\begin{itemize}
    \item \emph{Binary adaptation with constant effort}. In this case, an agent's willingness to take action is determined \emph{a priori}, and they will make the \emph{exact} changes to their features per the recommended recourse they receive.
    \item \emph{Continuous adaptation with flexible effort}. In this case, an agent's willingness to take action is determined by how much effort they need to make.  Further, how much they change their features is also probabilistic, and may result in them meeting, out-performing or under-performing the recourse recommendation.
\end{itemize}

\begin{figure*}[!t]
    \centering
    \includegraphics[width=0.7\linewidth]{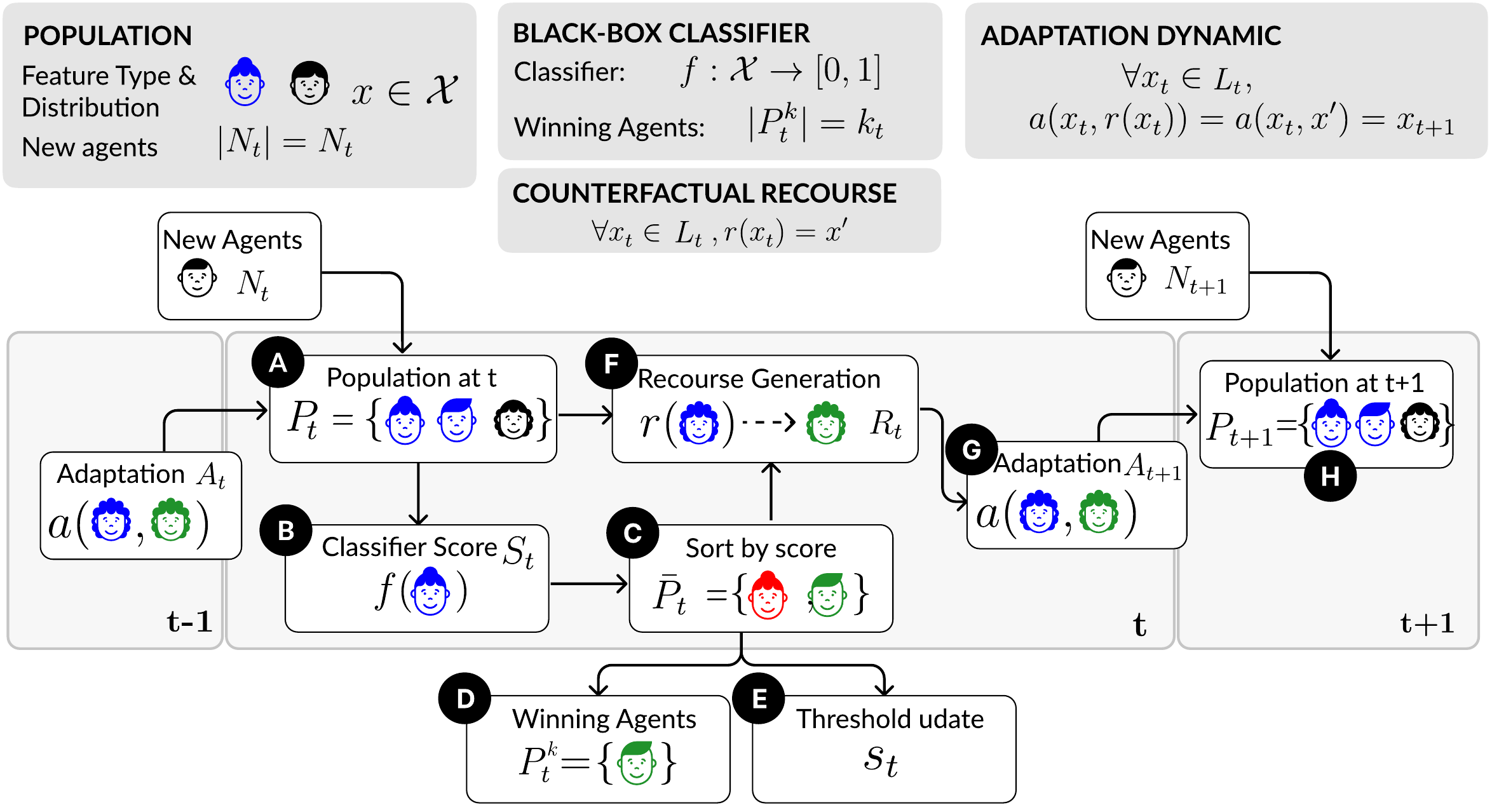}
    \caption{Overview of our simulation framework. \Circled{A} The initial population $P_t = A_t \cup N_t$ is the union of the agents  $A_{t}$ that had a negative outcome at time $t-1$ and new agents $N_{t}$ entering in the system. \Circled{B} The black-box classifier $f:\mathcal{X} \rightarrow [0,1]$ computes agents' scores $S_t$. \Circled{C} Agents are sorted according to the scores in $\bar{P}_t$. \Circled{D} The top-$k$ agents in $\bar{P}_t$ are selected as $P^k_t$ and exit from the system. \Circled{E} The threshold $s_t$ is updated to the lowest score of the agents in $P^k_t$. \Circled{F} For all the non-winning agents $L_{t}$, recourse is provided by $r:\mathcal{X} \rightarrow \mathcal{X}$. \Circled{G} The adaptation function $a:\mathcal{X}\times \mathcal{X} \rightarrow \mathcal{X}$ generates the evolution of each agent $x_t \in L_{t}$ from time $t$ to $t+1$, given the recourse $r(x_t)=x'$, with the final output $a(x_t,r(x_t))=a(x_t,x')=x_{t+1}$. \Circled{H} The initial population at $t+1$ is $P_{t+1} = A_{t+1} \cup N_{t+1}$.}
    \label{fig:framework}
\end{figure*}

\subsection{Simulation Framework}
\label{sec:simulation_framework}

We consider an environment where agents compete for access to a scarce resource repeatedly, at discrete time steps $t_0, \ldots, t_n$. The number of resources at each time step is given by $k$, which determines the number of positive outcomes (\eg loans) that can be assigned at each time step. The simulation begins with an initial population of agents $P_0$, all with synthetically-generated features.

At each time step, every agent receives a score from a machine learning classifier $f:\mathcal{X} \rightarrow [0,1]$, representing their attempt at a positive outcome (\eg applying for a loan). The agents with the top-$k$ scores at each time step receive a positive outcome and exit the environment (\eg they receive a loan). Agents that receive a negative outcome are given recourse recommendations from a function $r:\mathcal{X} \rightarrow \mathcal{X}$, and have the chance to act on those recommendations at a later time step. The likelihood an agent will ``take action'' and change their features is governed by an ``agent behavior'' function $a_{l,g} : \mathcal{X} \times \mathcal{X} \rightarrow \mathcal{X}$. Note that both the likelihood of taking action (or not) \emph{and} the amount action taken are governed by $a_{l,g}$.

We encode $l$ and $g$ as described in Section~\ref{sec:problem:simulation} as hyperparemters for the adaptation function $a_{l,g}$. Recall that $g \in [0,1]$ determines the global difficulty of achieving recourse and is set \emph{a priori} for the simulation. We let $l \in [0,1]$ be defined separately for each agent when they enter the simulation, and it reflects each agent's individual willingness to take action. We draw $l$ from a random distribution, and a higher value of $l$ means that an agent is more willing to act upon a recourse recommendation. Note that $l$ is mutable from one time-step to another.  For example, under the \emph{flexible} setting (see Figure~\ref{fig:adaptation-vs-effort}), if an agent finds themselves closer to the score threshold, their willingness to take action may increase.

Importantly, the population of agents is not fixed: at each time step,  new agents (\ie new loan applicants) join the environment. We model these dynamics by representing the population of agents over time as a sequence  $P = \{P_{t_0}, \dots, P_{t_T}\}$. At the end of the simulation, we retrieve $P$ that contains the final state of each agent. 

\renewcommand{\algorithmicrequire}{\textbf{Inputs:}}
\renewcommand{\algorithmicensure}{\textbf{Output:}}

\begin{algorithm}[t]
    \caption{Simulating multi-agent recourse over time}
    \label{alg:recourse_over_time}
    \begin{algorithmic}[1]
    \Require Classifier $f:\mathcal{X} \rightarrow [0,1]$; function for generating recourse recommendations $r:\mathcal{X} \rightarrow \mathcal{X}$;  function describing agent recourse actions $a_{l,g}:\mathcal{X} \times \mathcal{X} \rightarrow \mathcal{X}$; global difficulty setting $g$; agent-level willingness $l$ for each agent; number of time steps $T$; number of favorable outcomes per time step $k$; initial population size $p$; number of new agents at each time step $n$
    \Ensure The sequence of sets of agents $\{P_{t_0}, \dots, P_{t_T}\}$ in their final state for each time step.
    
    \State $t \gets 0$
    \State $P_0 \gets $ create the initial population of $p$ agents 
    \State $A_t = \emptyset$ 
    \While{$t<T$}
        \If{$t \neq 0$} 
            \State $N_t \gets$ set of $n$ agents joining the simulation;
            \State $P_t \gets A_t \cup N_t$, full population of agents;  
        \EndIf
        
        \State $S_t \gets \{x_t, s | \forall x_t \in P_t, s = f(x_t)\}$, compute agent scores;  
        \State $\bar{P}_t \gets sort(S_t)$, sort agents by score;  
        \State $P^k_t \gets select\_top\_k(\bar{P}_t)$, select $k$ highest-scoring agents, assign them a positive outcome; 
        \State $s_t \gets update(P^k_t)$, update the decision threshold to be the minimum score from $P^k_t$;  
        \State $L_t \gets P_t \setminus P^k_t$, select agents who did not receive a positive outcome;
        \State $R_t \gets \{x'| \forall x \in L_t, x'= r(x)\}$, generate recourse recommendations;  
        \State $A_{t+1} \gets \{x_{new}| \forall x \in L_t, x_{new} = a_{l_x,g}(x,x')\}$, for each agent, determine what action they take (if any);  
        \State $t \gets t + 1$, increase time-step
    \EndWhile
    \end{algorithmic}
\end{algorithm}

\subsection{Quantifying the Reliability of Recourse}
\label{sec:problem:metrics}

As discussed in Sections~\ref{sec:introduction} and~\ref{sec:related}, prior work on time-related effects of recourse is very limited.  To the best of our knowledge, only two  recourse reliability metrics have been defined to-date. Ferrario and Loi propose measuring the reliability of recourse using \emph{``Unfortunate Counterfactual Events (UCEs)''} that occur when a counterfactual explanation used as a recourse recommendation becomes invalidated at a later time due to updates in the underlying machine learning model~\cite{ferrario2022robustness}. Pawelczyk \etal define a metric called \emph{Recourse Invalidation Rate} that quantifies the probability that a recourse recommendation becomes invalidated for a single individual due to changes in the way recourse is implemented by that individual~\cite{pawelczyk2022probabilistically}.

In this paper, we propose an alternate metric that quantifies how well recourse recommendations meet individuals' expectations. In other words, we offer a metric that quantifies whether the ``promise'' of recourse matches reality. We call this metric \textbf{recourse reliability,} and see it as an important contribution to the literature, for two reasons: first, unlike previous metrics, it quantifies \emph{system-level} behavior, as opposed to focusing on recourse for a single individual. Second, our metric is not tied to counterfactual explanations, and is agnostic to the way the underlying recourse recommendation is generated. For example, it is can be used to quantify reliability of \emph{principle reason explanations}~\cite{barocas2020hidden}.

\smallskip
\noindent \textbf{The impact of competitive effects on the score threshold for a positive outcome.} Consider the simulation framework described in Section~\ref{sec:simulation_framework}. At each time step $t > 0$, an agent that received a negative outcome in the previous time step has a chance of altering their feature space in such a way that it either meets or exceeds the previous score threshold for a positive outcome $s_t$. Typically, an agent that carries out this behavior would expect to receive a positive outcome at the next time step, $t+1$. However, due to competitive effects (described in Figure~\ref{fig:sketch}) the agent may not receive a positive outcome. Specifically, we consider two competitive effects:
\begin{enumerate}
    \item Agents already present in the environment may act on recourse recommendations and exceed the score of other agents.
    \item New agents entering the environment may have scores that exceed the score of other agents.
\end{enumerate}

In both  cases, it is possible that the score threshold for a positive outcome increases so that $s_{t+1} > s_t$.  Intuitively, the score threshold can only remain constant if the number of agents acting on recourse and the number of agents is equal to the resource constraint $k$. This is shown in the following equality:

\begin{equation}\label{eq:ineq}
\frac{\mathbb{E}[\sum_{x_t \in P_t} 1_{[f(x_t)>s_{t-1}]}]}{k}=1
\end{equation}

Importantly, if the equality \emph{does not hold}, then the score threshold for a positive outcome may increase at future time steps. As a result, the reliability of algorithmic recourse recommendations cannot be guaranteed.

\smallskip
\noindent \textbf{Defining recourse reliability.}
Let us denote the set of agents that changed their features per a recourse recommendation and \emph{met or exceeded the score threshold $s_{t-1}$} as $C_t = \{x_t \in A_t| f(x_t)\geq s_{t-1}\}$. The agents $C_t$ successfully acted on a recourse recommendation and thus \emph{expect to receive a positive outcome}.  Recall $P^{k_t}$ is the set of agents that received a positive outcome at time step $t$. Using this notation, we can define \textbf{recourse reliability at time t} as:

\begin{equation}
    RR_t = \frac{|C_{t} \cap P^k_t|}{|C_{t}|}
    \label{eq:rr}
\end{equation}

Stated plainly, recourse reliability at time $t$ is the proportion of agents who acted on recourse and \emph{received a positive outcome}, out of all those agents who acted on recourse and \emph{expected a positive outcome}. In this way, recourse reliability is a measure of how well recourse expectations are met for agents.

%% file: experiments.tex
\section{Empirical Analysis}
\label{sec:methods}

To better understand the behavior of recourse reliability, we conducted extensive empirical analysis using the simulation framework described in Section~\ref{sec:simulation_framework}. Our analyses also allow us to demonstrate how the various concepts defined previously like adaptation, effort, competitive effects impact algorithmic recourse. The following analyses were executed using Python with its standard libraries.
Our implementation and all supporting code (including the empirical analysis reported here) are publicly available on GitHub\footnote{https://github.com/joaopfonseca/recourse-game/}. 

We report results of simulations using the parameters summarized in Table~\ref{tab:parameters}. All reported results represent 20 executions with different initial random seeds, saved for reproducibility.

\subsection{Experimental Setup}
 
Table~\ref{tab:parameters} summarizes the parameters of the simulations used in our experimental evaluation. We refer to Figure~\ref{fig:framework} for a visual summary of the components of the framework, and their interactions.

\begin{table}[t!]
    \centering
     \caption{Summary of simulation parameters}
    \begin{tabular}{p{1cm}p{3.5cm}p{2.5cm}}
        \toprule
        Symbol & Parameter & Settings\\
        \midrule
        $p$ & Initial number of agents & 100 \\
        $k$ & Number of favorable outcomes per time step & 10\\
        $n$ & Number of new agents per time step  & $\{0.8k, 0.9k, k, 1.1k, 1.2k\}$\\
        $a_{l,g}(x, x')$ & Describes \emph{adaptation} and \emph{effort} of agents & \{binary, continuous\} $\times$ \{constant, flexible\} \\
        $l$ & Agent-level (local) willingness of acting on recourse&  $[0,1]$\\
        $g$ & Global ease of acting on recourse & $[0,1]$\\
        T & Number of time steps & 50 \\
        \bottomrule    
    \end{tabular}
    \label{tab:parameters}
\end{table}

\paragraph{Population} Agents are generated over a 2-dimensional feature space, sampled independently at random.  Each feature is sampled from a Gaussian distribution, where $x = (x_1,x_2)$ and $x_i \sim \mathcal{N}(\mu = 0.5, \sigma = 0.\overline{3}), i=1,2$. All simulations have an initial population $p = 100$ agents. Of these, $k=10$ receive the favorable outcome at every time step and exit the system.  New agents $n$ enter the simulation at every time step, where the number of new agents is fixed per simulation, and varies between 8 and 12 across simulations.

\paragraph{Classifier} The framework described in Section~\ref{sec:simulation_framework} is model-agnostic, but for our experiments we used a simple logistic regression classifier to determine the score for each agent at each time step. The target variable $y_i$ for the prediction task is created randomly using a binomial distribution. 

\paragraph{Calculating recourse recommendations.} Our simulation framework is also agnostic to the method used for generating recourse recommendations for each agent. We use a simple approach to generate single-agent counterfactual explanations for linear classification, based on the work by~\citet{ustun2019actionable} and its open source implementation.

%% file: results.tex
\subsection{Experimental Results}
\label{sec:experiments:results}

We present our results based on the different types of agents behavior, and show full results for three parameter settings: 
\begin{enumerate}
    \item Binary adaptation with constant effort.
    \item Continuous adaption with constant effort.
    \item Continuous adaption with flexible effort
\end{enumerate}

We do not report results for the binary adaptation flexible effort setting because it does not contain unique insights as compared to the other settings. Further, we do not explore the effect of varying $l$ within each experiment (\ie agent leveling willingness) but instead define $l$ \emph{a priori} for each setting. We do this because we are primarily interested in the impact of the (global) difficulty of recourse and of the number of new applicants at each time-step.  

\smallskip
\subsubsection{Binary adaptation with constant effort}~\label{sec:binary}

Recall that binary adaptation means that all agents $x$ will adapt to match the counterfactual configuration $x'$ based on some fixed probability. We express this probability in the following way:

\begin{equation}
    a_{l,g}(x_t, x_t') = (1-l) \times x_t + l \times x_t'
\end{equation}

\begin{figure}
    \centering
    \includegraphics[width=1.0\linewidth]{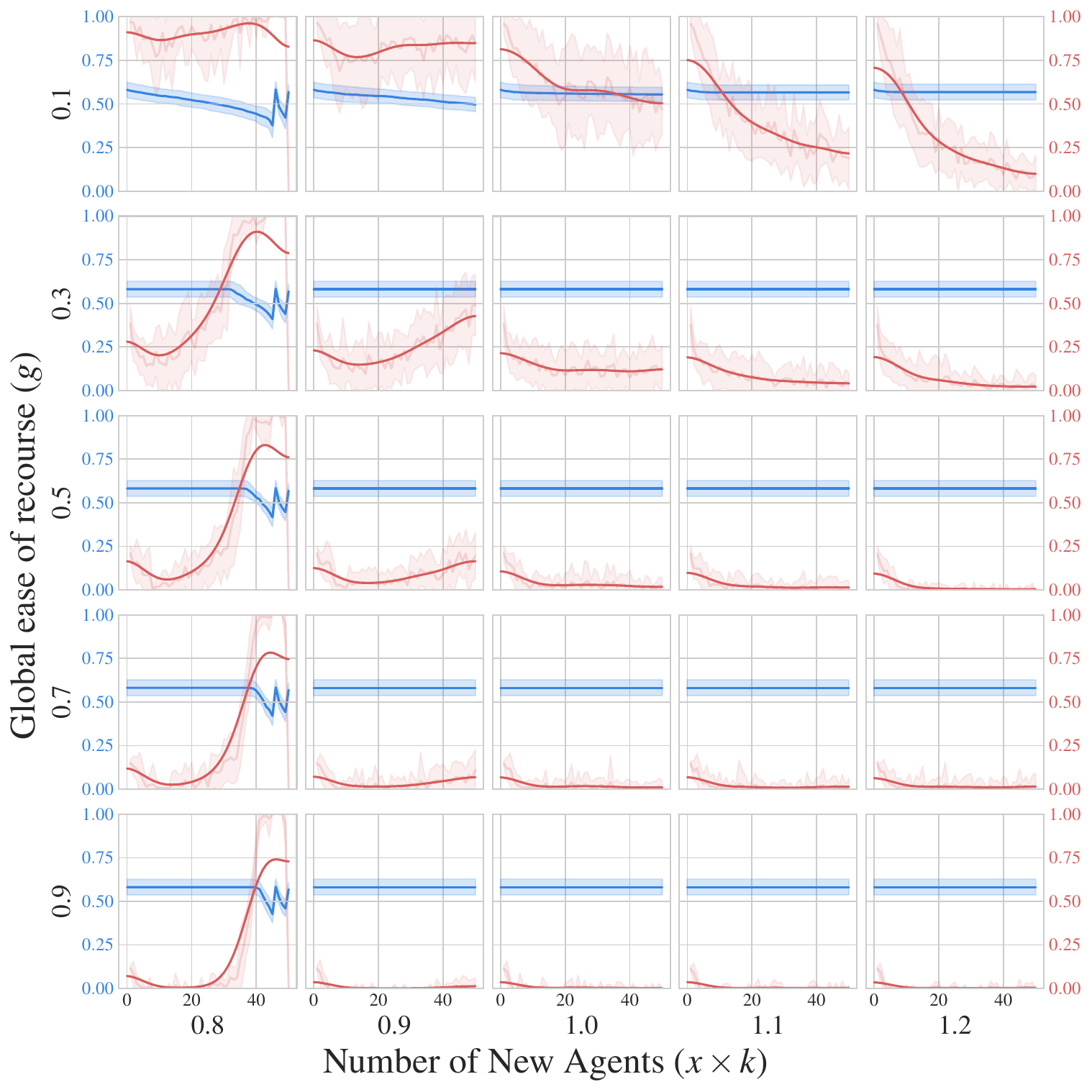}
    \caption{Binary adaptation with constant effort.  Score threshold values for a positive outcome (blue lines) and recourse reliability (red lines) over 50 time steps (error bars given by running 20 simulations with different random seeds). The $y-$axis in each subgraph is the score of individual agents, and the $x-$axis in each subgraph is the time step. The large $y-$axis is the difficulty of acting on recourse $g$, and the large $x-$axis is the number of new agents at each time step $N_t$.}
    \label{fig:multiple-runs-binary}
\end{figure}

We sample $l$ from a Bernoulli distribution where $l \sim Bernoulli(g)$. Figure \ref{fig:multiple-runs-binary} shows the score threshold for a positive outcome $s_t$ and the recourse reliability $RR_t$ over 50 time-steps, varying the number of agents taking recourse action and the number of new agents entering at each time step. There are several observations that can be gleaned from the figure. First, over nearly all settings, the threshold $s_t$ is relatively constant, with imperceptible drifts over time. Second, and significantly, the recourse reliability $RR_t$ always tends to decrease (at different rates). Third, in settings where the $g$ is approximately $0.5$ or greater and the number of new agents is approximately $0.9 \times k$ or greater, the $RR_t$ is close to $0$. This is because with so many agents adapting, the environment becomes incredibly competitive and it is difficult for agents to achieve a successful recourse event.

\smallskip
\subsubsection{Continuous adaptation with constant effort}~\label{sec:innate}

Recall that constant effort is based on the idea that each agent has an \emph{a priori} willingness to act on a recourse recommendation, and it does not change over time.  We sample $l$ a random uniform distribution where $l \sim \mathcal{U}(0,1)$, and as described earlier, $g$ is an input parameter of the simulation. 

\begin{figure}
    \centering
    \includegraphics[width=1.0\linewidth]{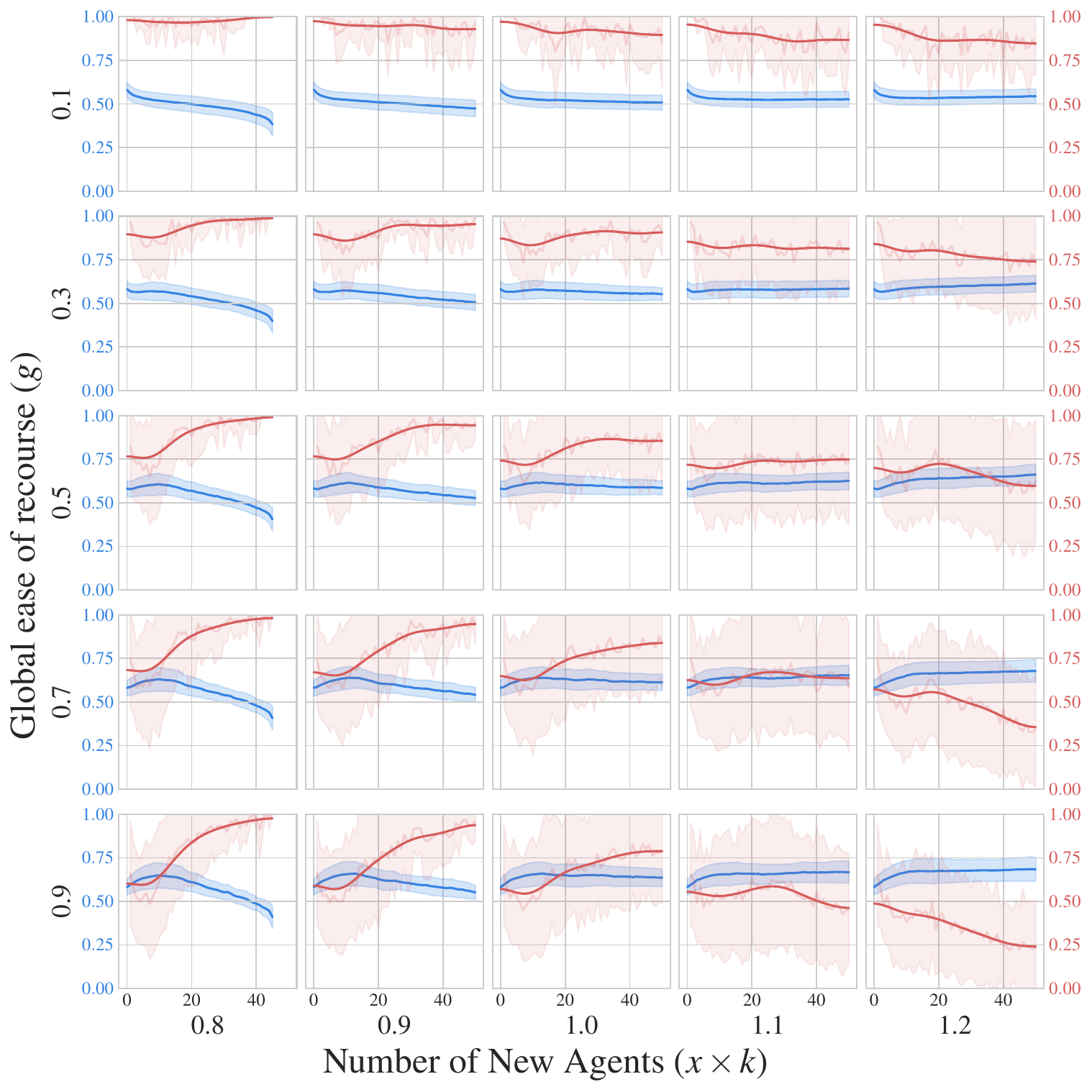}
    \caption{Continuous adaptation with constant effort. Identical details as in Figure~\ref{fig:multiple-runs-binary}.}
    \label{fig:multiple-runs-willingness}
\end{figure}

Figure~\ref{fig:multiple-runs-willingness} is analogous to Figure~\ref{fig:multiple-runs-binary}. Notice that in this case, the volatility of the recourse reliability score $RR_t$ varies with a more discernible pattern: although the score threshold for a positive outcome $s_t$ also varies with both parameters, it always decays when the number of new agents (8 or 9) is lower than the number of favorable outcomes (10). Overall, the recourse reliability scores follow non-linear trajectories with high volatility.

\smallskip
\subsubsection{Continuous adaptation with flexible effort}~\label{sec:adaptive}

Recall that flexible effort means that an agent's willingness to take action on recourse is based on the agent's distance to the score threshold for a favorable outcome. It follows the assumption that an agent with a score closer to the score threshold will have a greater incentive to exert effort as compared to an agent with a lower score. In this case, we define the agent level willingness $l$ as follows:

\begin{equation}\label{eq:l}
    \begin{gathered}
        l = \frac{1}{s_t - f(x)} \\
    \end{gathered}
\end{equation}

\begin{figure}
    \centering
    \includegraphics[width=1.0\linewidth]{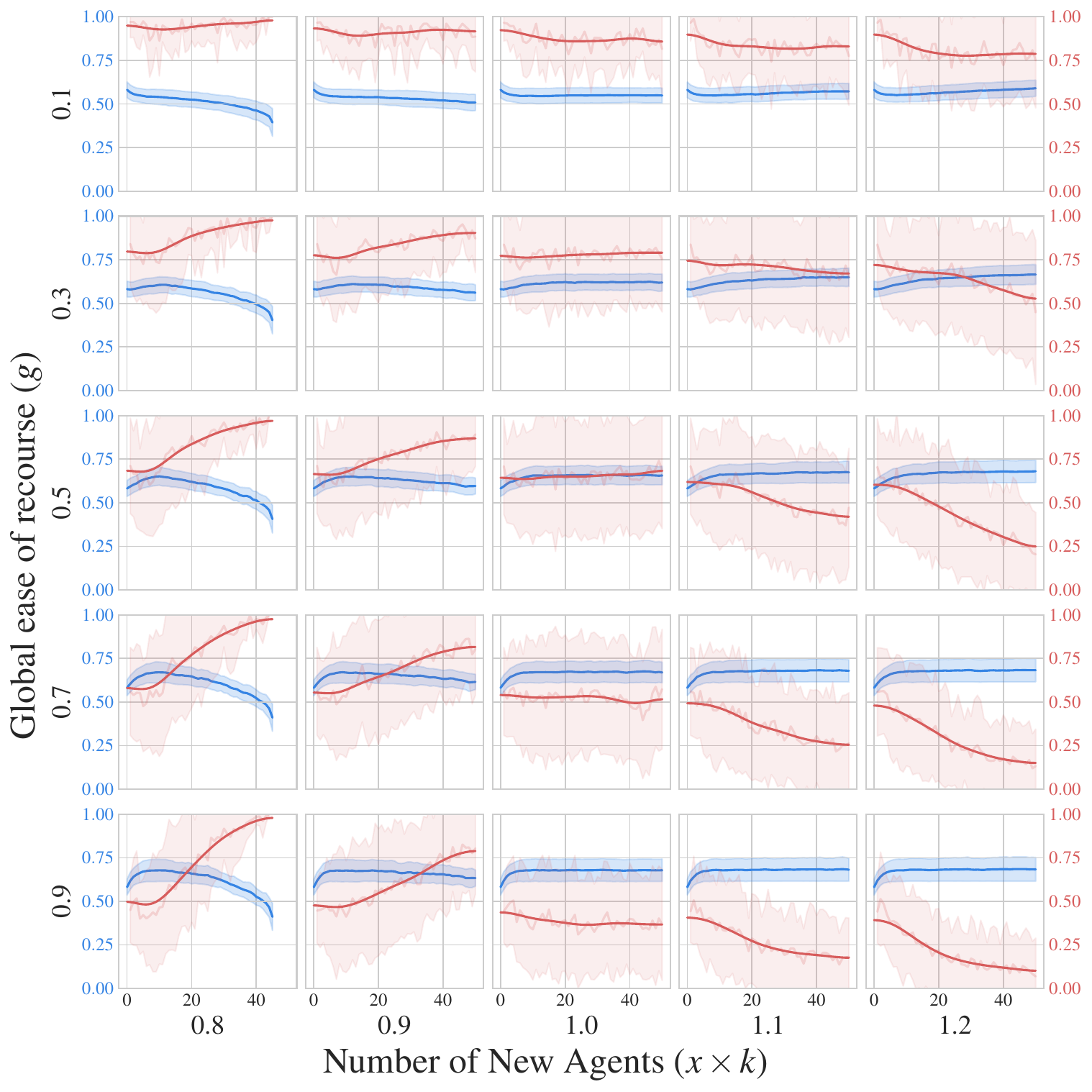}
    \caption{Continuous adaptation with flexible effort. Identical details as in Figure~\ref{fig:multiple-runs-binary}.}
    \label{fig:multiple-runs-gaussian}
\end{figure}

Figure \ref{fig:multiple-runs-gaussian} is analogous to Figure~\ref{fig:multiple-runs-binary}. Although one observes lower volatility of both the score threshold and the recourse reliability as compared to the continuous adaptation with constant effort setting (Section~\ref{sec:innate}), some parameter settings yield high volatility. Specifically, larger $g$ and lower $N_t$ lead to lower recourse reliability. 

\smallskip
\subsubsection{Summary of experimental results}
\label{sec:results:summary}

Generally, we observe that recourse reliability will decrease as the global difficulty $g$ increases (\ie it is harder for agents to act on recourse), and as the number of agents increases above the resource constraint $k$. Intuitively, this reflects the idea that a more competitive environment will result in lower recourse reliability. 

\emph{Adaptation.} In terms of agent behavior, holding all other parameters constant, binary adaptation has a negative impact on recourse reliability overall as compared to continuous adaptation. Again, our intuition is that binary adaptation models a more competitive environment since agent's are ``quicker'' to implement recourse. Also, in this setting, agents often find themselves having their score tied with others since all agents are changing their score to \emph{exactly} match the threshold.

\emph{Effort.} In a similar fashion, flexible effort has a more negative impact on recourse reliability as compared to constant effort, likely due to making more the environment more competitive. Under flexible effort, agents that are closer to the threshold will act on recourse and pass the threshold more frequently.

%% file: discussion.tex
\section{Discussion}
\label{sec:disc}

In this paper, we developed a framework that simulates recourse in a \emph{multi-agent environment over time}.  In the experiments conducted using our framework, we found that, in the vast majority of cases, the score threshold for a positive outcome either decreases or increases over time.  Both of these cases can be harmful to the agents: If the threshold increases over time, then recourse reliability suffers (\ie some of the agents that were promised a positive outcome will not receive one even if they reach the threshold).  If the threshold decreases over time, then recourse recommendations are reliable, but the effort of agents acting on recourse may be wasted (\ie they worked harder than necessary to achieve the positive outcome).  

Importantly, \emph{across all of our simulations}, we found that the score threshold rarely remains stable over time, and, further, that threshold stability, which is desirable from the point of view of the agents, is the most difficult to parameterize for.

Our observations have a significant socio-technical implication: \textbf{In systems that administer recourse \emph{without considering possible changes  over time}, individuals are at best being given unrealistic expectations about what will happen to them if they follow recommendations for recourse, and, at worst, the system designers are being irresponsible ---and possibly damaging--- by falsely promising a reward to individuals for their efforts.}  This finding supports concerns by Barocas~\etal and other authors about the need to study recourse under the assumption that the system will change over time~\cite{barocas2020hidden,ferrario2022robustness,pawelczyk2020counterfactual}.

For a real-world example, let us consider college admissions. There is substantial evidence that the selectivity of US universities changes over time~\cite{hoxby2009changing}. One can imagine a scenario where an applicant is denied admission to their top-choice university in one year, and is recommended a set of improvements to make to their application (\eg increase their SAT score, increase the number of volunteering hours).  However, these improvements may turn out to be insufficient for admission when the individual re-applies to the same university one or two years later.

\smallskip
\noindent \textbf{Guidance for system-level decision-makers.}    Our framework can be used to provide guidance to system-level decision-makers like banks, colleges, and governments.  The parameters of our framework --- the number of new individuals at each time step, the number of individuals that re-apply at each time step, and the expected level of improvement among the individuals who re-apply--- can all be easily measured in practice. Once these parameters are known, our framework can anticipate changes in the threshold over time, and provide insight regarding the time-step at which specific changes are expected to occur.  Another use case for our framework is that, if any of these values can only be estimated rather than measured, one could run many experiments sweeping over a broad range of values and scenarios to better understand possible outcomes.

There are two concrete actions that decision-makers can take based on empirical insights. First, recourse reliability can be measured and used to provide an uncertainty estimate to  individuals undertaking recourse action. For example, the system could generate a  recommendation in the following way: ``If you make the following $X$ changes by time $t$, then there is a $Y\%$ chance that you will receive a positive outcome.''   Second, if it is possible to adjust resource constraints (\eg the number of loans being given, or the number of spots in an incoming college class), our framework can inform what setting of the resource constraint maximizes recourse reliability. For example, colleges could estimate the ideal incoming class-size that achieves a stable threshold for admissions.

%% file: conc.tex
\section{Conclusions, limitations, and future work}
\label{sec:conc}

In this paper, we sought to close a significant gap in the literature, and conducted multi-agent, multi-time step analysis of algorithmic recourse within competitive environments.  We developed a simulation framework that allows the configuration of a diverse set of components, such as incorporating external interventions/shocks, problem-specific adaptation functions, introduction of new agents and resource constraints, and opens up opportunities to study how these settings affect the reliability of algorithmic recourse over time.  The software implementation of our framework is available in a public GitHub repository. The major finding of our work is that recourse is only reliable under a very specific set of conditions, leading to an important insight for people who design recourse methods, and for people who receive recourse recommendations.  It is our hope that this paper will lead to more robust and reliable algorithmic recourse methods. 

\smallskip
\noindent \textbf{Limitations and future work.} In this paper, we presented an agent-based framework that has undergone rigorous testing using \emph{simulated data}. Our future plans involve evaluating the practical applicability of our insights by incorporating real-world data and deployment scenarios. 

A further limitation of our work is that our adaptation functions attempt to model human behaviour, which is complex and not always rational.  In future work, we plan to further validate these functions, and to develop methods that learn adaptation behavior from historical data.

Further future work involves designing additional metrics, and exploring the impact of different feature distributions, and distributions of adaptation and effort, on recourse reliability, efficiency, and fairness.

%% file: ack.tex
\begin{acks}
This research was supported in part by NSF Awards No. 1916505 and 192265, by the NSF Graduate Research Fellowship under Award No. DGE-2234660, by research grants from the Portuguese Foundation for Science and Technology (``Fundaç\~{a}o para a Ci\^{e}ncia e a Tecnologia'') references SFRH/BD/151473/2021 and UIDB/04152/2020, and by the New York University Center for Responsible AI.
\end{acks}